\documentclass[pdflatex,sn-nature]{sn-jnl}

\usepackage{graphicx}
\usepackage{amsmath,amssymb}
\usepackage{booktabs}
\usepackage{xcolor}
\usepackage{url}

\begin{document}

\title[Representing contradiction in Gemma 3]{When Language Describes the Impossible: Internal Representations of Contradiction and Paradox in Gemma 3}

\author*[1]{\fnm{Yoon Pyo} \sur{Lee}}\email{yoonpyo2@illinois.edu}
\affil*[1]{\orgdiv{Department of Nuclear, Plasma, and Radiological Engineering}, \orgname{University of Illinois Urbana-Champaign}, \orgaddress{\city{Urbana}, \state{Illinois}, \country{USA}}}

\abstract{Language can describe states of affairs that are contradictory, paradoxical, or otherwise difficult to make coherent. Whether a language model treats these cases through a shared internal representation or only through surface associations remains unclear. I report an exploratory activation study of Gemma 3 4B IT using 85 controlled stimuli derived from 17 philosophical seed cases. The model classified each statement as coherent, contradictory, paradoxical, or underdetermined while residual-stream states were recorded immediately before answer generation. Exact four-way behavioral accuracy was 55.3\%, with a marked tendency to call heterogeneous cases paradoxes. Nevertheless, group-held-out linear probes distinguished coherent from noncoherent stimuli with a peak balanced accuracy of 0.79 at layer 18, compared with 0.54 for a token-count baseline and at most 0.52 for TF--IDF controls (family-restricted permutation, Bonferroni-adjusted $P=0.018$). A pretrained Gemma Scope 2 sparse autoencoder further identified candidate features with differential activity across the four labels. These results show reproducible structure associated with the model's treatment of inconsistency, but do not establish a unitary concept of impossibility or a causal mechanism. The study is offered as a small empirical record of how philosophical contrasts appear in one transformer's activation space.}

\keywords{mechanistic interpretability, contradiction, paradox, sparse autoencoders, language models, philosophy of language}

\maketitle

\section{Introduction}\label{sec:introduction}

Language permits combinations that the world, a definition, or a system of rules does not. A speaker can say \emph{square circle}, \emph{married bachelor}, or \emph{this sentence is false} without first constructing a corresponding object or consistent state of affairs. The expressions are not defective in a single way. Some directly violate a definition, some assert both a proposition and its negation, some generate self-referential paradoxes, and others are grammatical while remaining semantically anomalous. Aristotle's formulation of the principle of non-contradiction gave one classical statement of the boundary: the same attribute cannot both belong and not belong to the same thing at the same time and in the same respect \cite{aristotle1984}. Much later, Wittgenstein placed the relation between proposition, possibility, and world at the center of the \emph{Tractatus} \cite{wittgenstein1922}. Chomsky's ``colorless green ideas'' demonstrated from a different direction that grammatical form can remain intact even when ordinary semantic composition becomes strained \cite{chomsky1957}. Such cases have remained philosophically attractive not merely as mistakes, but as boundary cases that expose where grammatical form, reference, truth conditions, and jointly satisfiable content come apart.

The question now also has practical stakes. Language-model output can be fluent and persuasive even when its claims are false, unsupported, or internally inconsistent. In a preregistered experiment, participants could not reliably distinguish GPT-3-generated tweets from human-written tweets, while the model produced disinformation that participants found more compelling than human-produced disinformation \cite{spitale2023}. A separate controlled study found that personalized GPT-4 opponents were more persuasive than human opponents in structured online debates \cite{salvi2025}. Falsehood, contradiction, and persuasion are not the same phenomenon: a false statement may be internally coherent, and an explicit contradiction need not be deceptive. Nevertheless, linguistic fluency can conceal failures of consistency from a reader. Whether a model internally registers such failures is therefore one part of understanding systems whose language can shape human belief.

These traditions do not supply a ready-made ontology for a transformer. A model trained on human language inherits both our ability to formulate impossible situations and the linguistic patterns through which we discuss them. Yet a transformer processes token sequences through high-dimensional continuous states \cite{vaswani2017}; it is not given an explicit inventory containing \emph{contradiction}, \emph{paradox}, or \emph{impossibility}. This leaves a narrow empirical question between the philosophical, the computational, and the practical: when a model encounters language that cannot coherently be the case, what changes inside it before it judges and explains that language?

The question must be posed cautiously. Layer-wise linear probes can test how readily labels are decoded from intermediate representations \cite{alain2016}, but decodability is not identity: a direction from which a label can be predicted is not thereby the philosophical concept named by that label. Nor does a successful probe show that the model causally uses the decoded information \cite{hewitt2019,belinkov2022}. Surface form can also produce impressive separability without revealing the computation of interest \cite{sahoo2026}. Sparse autoencoders (SAEs) offer a complementary decomposition of dense activations into sparsely active learned features, but their features likewise require validation and interpretation \cite{elhage2022,cunningham2023,lieberum2024}.

Here I treat philosophical contradictions and paradoxes as experimental stimuli rather than doctrines to be resolved. I construct controlled families containing a canonical statement, three surface transformations, and a nearby coherent control. I ask three descriptive questions: (i) how does a small instruction-tuned model behaviorally divide these cases; (ii) at what depth does coherence become linearly decodable when whole stimulus families are held out; and (iii) do pretrained SAE features show label-selective activity? The aim is deliberately modest. The experiment does not determine whether a model \emph{understands} impossibility. It records whether several human-defined kinds of inconsistency leave shared or distinguishable traces in one open-weight transformer. The deeper question is not whether an AI reproduces human distinctions, but whether an artificial system trained on human language can reveal the structural tensions between our descriptions of the world and the world itself.

\section{Results}\label{sec:results}

\subsection{The model recognized disturbance more readily than kind}

\begin{figure}[t]
\centering
\includegraphics[width=\textwidth]{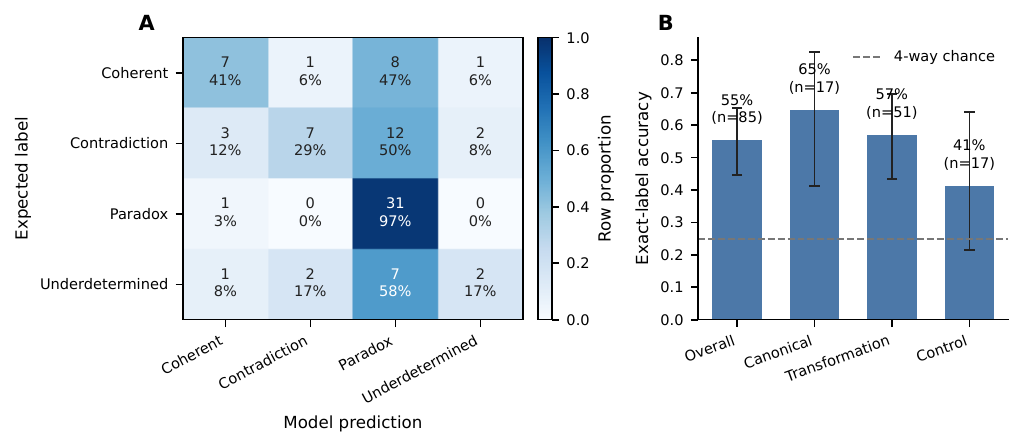}
\caption{\textbf{Behavioral classification of philosophical stimuli.} \textbf{a}, Row-normalized confusion matrix for Gemma 3 4B IT on 85 stimuli. Each cell reports count and row percentage. \textbf{b}, Exact four-way classification accuracy overall and by stimulus form; error bars are Wilson 95\% confidence intervals. The model achieved 55.3\% overall accuracy and showed a strong tendency to use the \emph{paradox} label.}
\label{fig:behavior}
\end{figure}

The stimulus set contained 85 prompts: 17 canonical cases, 51 transformations, and 17 coherent controls. Expected labels comprised 17 coherent, 24 contradiction, 32 paradox, and 12 underdetermined examples. Gemma 3 4B IT produced an exact expected label for 47 of 85 prompts (55.3\%). Accuracy was 64.7\% on canonical statements, 56.9\% on transformations, and 41.2\% on coherent controls (Fig.~\ref{fig:behavior}).

Errors were not symmetric. The model predicted \emph{paradox} for 58 of 85 prompts. It correctly identified 31 of 32 expected paradoxes, but also assigned that label to 8 of 17 coherent controls, 12 of 24 direct contradictions, and 7 of 12 underdetermined cases. Direct contradictions such as ``$P$ and not-$P$'' were therefore often recognized as problematic while being placed in the wrong subclass. When the four labels were collapsed to a binary coherent/noncoherent distinction, ordinary accuracy rose to 82.4\%, but balanced accuracy was only 0.67: recall was 92.6\% for noncoherent cases and 41.2\% for coherent cases. The model exhibited a broad rejection tendency rather than a balanced taxonomy of inconsistency.

\subsection{Coherence became linearly accessible across middle layers}

For each prompt, I extracted the residual-stream state at the final input token, immediately before the first answer token was generated. A separate regularized linear classifier was trained at each depth to distinguish expected coherent from expected noncoherent stimuli. Cross-validation was grouped by seed item: the canonical statement, its three transformations, and its coherent control always remained in the same fold. This prevents a probe from being trained on one wording of a case and tested on a nearby wording of the same case.

\begin{figure}[t]
\centering
\includegraphics[width=\textwidth]{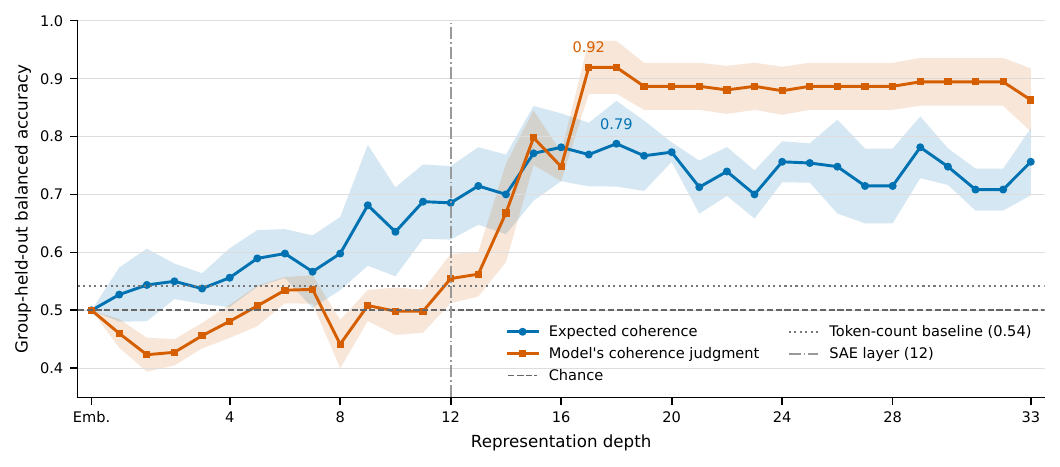}
\caption{\textbf{Layer-wise decodability of coherence.} Five-fold group-held-out linear probes distinguish coherent from noncoherent stimuli. Blue uses the expected binary label; orange uses the model's eventual binary judgment. Shading denotes the standard error across folds. The dotted baseline uses input token count alone. Expected coherence peaked at layer 18 (balanced accuracy 0.79), whereas the model's own judgment peaked at layer 17 (0.92).}
\label{fig:probe}
\end{figure}

At the embedding output, balanced accuracy was 0.50. Decodability increased gradually through the early and middle transformer blocks, reaching 0.69 at layer 12 and peaking at 0.79 at layer 18 (Fig.~\ref{fig:probe}). It remained above 0.70 through the final layer and ended at 0.76. Input token count reached 0.54 balanced accuracy, while word 1--2-gram and character 3--5-gram TF--IDF classifiers reached 0.49 and 0.52, respectively. At the observed peak, a 1,999-iteration permutation test that shuffled labels only within each five-example family gave $P=0.0005$; Bonferroni correction over the 35 tested depths gave $P=0.0175$.

A second probe used not the expected label but the model's eventual binary judgment. This target stayed near chance through much of the early network, rose abruptly in the middle-to-late layers, and peaked at 0.92 balanced accuracy at layer 17. This estimate is less stable than its magnitude suggests: only 12 of 85 model outputs were labeled coherent, and peak-layer fold scores ranged from 0.75 to 1.00. The temporal alignment is suggestive, but remains correlational; the probe was not used to intervene on generation.

\subsection{Low-dimensional geometry changed with depth}

\begin{figure}[t]
\centering
\includegraphics[width=\textwidth]{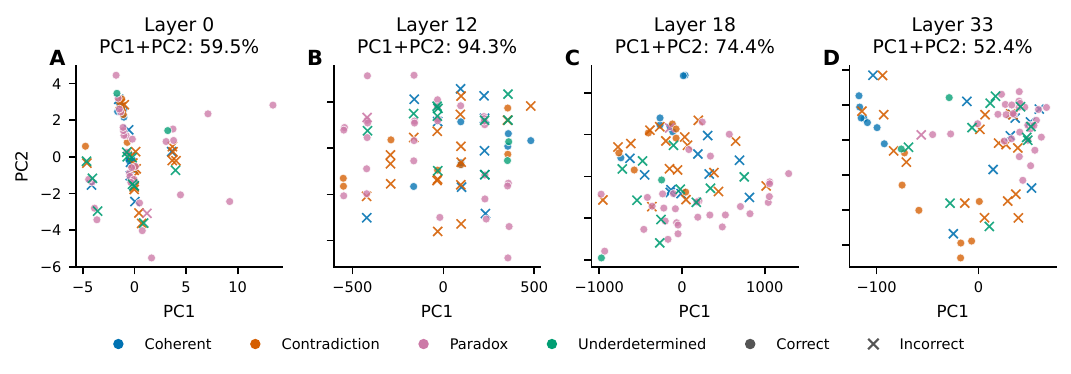}
\caption{\textbf{Evolution of residual-stream geometry.} Two-dimensional PCA of the final prompt-token representation at four depths. Color denotes the expected label; circles and crosses denote correct and incorrect four-way classifications. PCA was fitted independently at each layer and is descriptive rather than a test of separability.}
\label{fig:pca}
\end{figure}

Principal-component projections provide a descriptive view of the same residual states (Fig.~\ref{fig:pca}). The first two components explained 59.5\% of variance at layer 0, 94.3\% at layer 12, 74.4\% at layer 18, and 52.4\% at layer 33. At layer 12, PC1 alone explained 92.2\% and was almost perfectly correlated with residual-vector norm ($r=0.99995$); after unit normalization its explained variance fell to 26.1\%. The raw projection at this depth is therefore dominated by radial scale rather than label geometry. Across depths, the geometry reorganized substantially rather than separating the four expected labels into four stable clusters. Paradox examples became locally concentrated at several depths, whereas contradiction and underdetermined cases remained interspersed. Correct and incorrect model judgments also overlapped.

This lack of four clean clusters is consistent with the behavioral results. The model treated many direct contradictions and underdetermined puzzles as paradoxes; the residual geometry need not respect a sharper human taxonomy than the model's output does. Conversely, the linear-probe result shows that the absence of visually isolated clusters does not imply the absence of decodable coherence information. PCA preserves variance, not necessarily the directions most useful for classification.

\subsection{Sparse features showed class-selective activation patterns}

I next encoded layer-12 residual states with the official Gemma Scope 2 JumpReLU SAE for Gemma 3 4B IT \cite{mcdougall2025}. The SAE maps the 2,560-dimensional residual state to 16,384 non-negative features. Across the philosophical prompts, a mean of 10.9 features was active per prompt, close to the intended sparse regime of this checkpoint.

\begin{figure}[t]
\centering
\includegraphics[width=\textwidth]{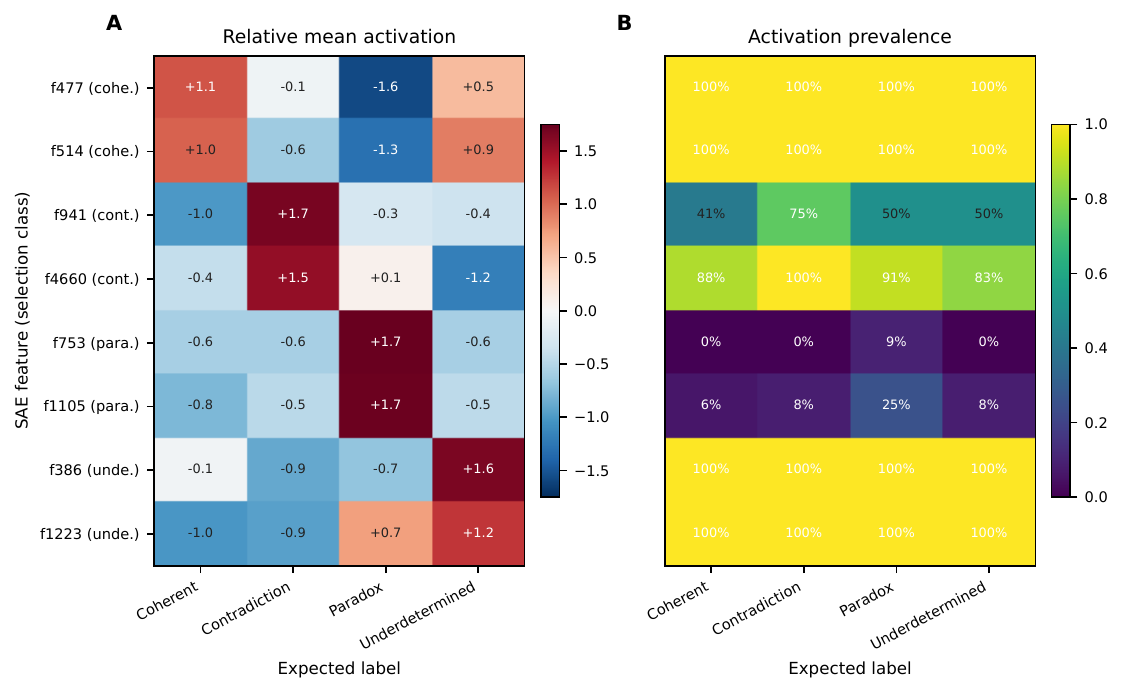}
\caption{\textbf{Label-selective Gemma Scope 2 SAE features.} Exploratory sparse-feature analysis at layer 12 using the official 16k-feature residual-stream SAE. Two candidate features were selected per expected class. \textbf{a}, Mean log-activation standardized within each feature across classes. \textbf{b}, Fraction of prompts on which each feature was active. Feature selection and visualization use the same small dataset; these features are correlational candidates, not established semantic detectors.}
\label{fig:sae}
\end{figure}

For an exploratory contrast, I selected two candidate features per expected label. A candidate had to activate on at least 5\% of that label's prompts, have its highest mean log-activation in that label, and rank highly by its within-feature standardized class mean. Several features showed differential magnitude despite being active for nearly every prompt (for example, features 477 and 514 for coherent cases and 941 and 4660 for contradictions). Two paradox-associated candidates were sparser: feature 753 fired on 9\% of paradox prompts and on none of the other classes, while feature 1105 fired on 25\% of paradox prompts and 6--8\% elsewhere (Fig.~\ref{fig:sae}). Features 386 and 1223 had their largest mean activation for underdetermined cases.

These are candidates, not semantic identifications. Feature selection and visualization used the same 85 examples, no independent corpus was used to assign natural-language descriptions, and high specificity can be produced by a small number of surface patterns. Several dense candidates also correlated strongly with layer-12 residual norm (absolute $r$ up to 0.91), so some magnitude contrasts may reflect overall activation scale. The SAE analysis therefore narrows possible sites for follow-up inspection but does not establish ``paradox neurons.''

\section{Discussion}\label{sec:discussion}

\subsection{What, if anything, has the model represented?}

The narrow empirical answer is that information associated with coherence is present in the model's residual stream before it generates an explicit classification. That information is weak or absent at the input embedding, grows across depth, survives held-out stimulus families, and becomes especially predictive of the model's own decision in layers 17--18. It is also not a proxy for surface form: length and lexical baselines remained near chance, and the family-stratified permutation test kept the peak significant after correction across depths. The SAE analysis supplies a second description in which a small set of learned features changes activation across the human labels.

The tempting stronger answer is that the model has represented \emph{impossibility}. I do not think the present experiment warrants it. First, the target categories are themselves interpretations imposed by the dataset. A liar sentence, a square circle, and the Ship of Theseus do not fail in the same manner. Second, the model often collapses those distinctions into \emph{paradox}. Third, a linearly decodable property may be correlated with a decision without taking part in the computation that causes it. Finally, an SAE feature is a learned coordinate system for activation space, not a guarantee that the coordinate coincides with a human concept.

There is nevertheless a philosophically interesting asymmetry in the result. The model's errors suggest that it more readily represents a broad disturbance---something like ``this statement resists an ordinary coherent reading''---than the particular source of that disturbance. This could reflect instruction-tuning data in which \emph{paradox} functions as a general conversational response to puzzling claims. It could also reflect a genuinely shared internal signal for violated expectations. Distinguishing the two requires the causal interventions outlined below.

Wittgenstein's early picture of a proposition ties sense to a possible configuration of objects \cite{wittgenstein1922}. A transformer complicates that picture: the model has no directly inspected world against which a sentence is compared, only learned parameters, current context, and internal state. Physical or logical ``possibility'' for such a system may therefore amount, operationally, to compatibility with constraints sedimented in language and activated in context. That is not the same as a metaphysical space of possible worlds. It is also more than a word list if the signal generalizes beyond phrasing. The group-held-out probe and the near-chance lexical baselines are evidence for the latter; the former is a question no activation study settles.

This motivates a restrained formulation. The experiment identifies a reproducible internal structure that covaries with, and predicts, the model's treatment of coherent and noncoherent language. Whether that structure deserves the name \emph{concept} depends on further criteria: generalization to new rule systems, robustness across domains and models, and causal participation in judgment.

\subsection{Limitations and next experiments}

The study is intentionally small. It uses one instruction-tuned 4-billion-parameter model, 17 seed families, English prompts, one prompt template, one sampled position, and one SAE layer. Moreover, every coherent example is a control and every canonical or transformed target is noncoherent, so expected coherence is confounded with stimulus role; the near-chance TF--IDF controls reduce but do not eliminate this concern. Although grouping prevents transformation leakage, entire categories are represented by few conceptual families; the three underdetermined families are especially sparse. Confidence intervals across prompts therefore overstate independence because transformations share a seed.

The labels also combine philosophical analysis with pragmatic annotation. For example, the Epimenides-style statement is underdetermined rather than strictly contradictory without additional assumptions, and the Ship of Theseus poses an identity question rather than an impossible state. Such distinctions are part of the object of study, but they also make a four-way accuracy number dependent on the chosen taxonomy.

Linear probing establishes accessibility, not use. Future work should use role-balanced minimal pairs, artificial worlds whose rules are specified in the prompt, and cross-domain transfer. The most informative causal extension would intervene on candidate residual directions or SAE features and measure selective changes in classification while monitoring unrelated language behavior. SAE candidates should be evaluated on a much larger independent corpus, with top-activating contexts inspected before assigning semantic descriptions. Recent work showing that probes or SAE contrasts can track task format and lexical artifacts rather than reasoning makes these controls essential \cite{sahoo2026,ma2026}.

\section{Methods}\label{sec:methods}

\subsection{Stimulus construction}

Seventeen seed cases were assembled from familiar logical, philosophical, and linguistic examples. Table~\ref{tab:stimuli} gives representative families. Each family contained one canonical English statement, three transformations (two paraphrastic declaratives and one question where possible), and one nearby coherent control. This produced 85 prompts. The transformations were written to preserve the underlying case while varying surface form; controls were written to preserve topic vocabulary where possible while removing the inconsistency.

\begin{table}[t]
\caption{Representative stimulus families.}\label{tab:stimuli}
\centering
\begin{tabular}{@{}p{0.18\textwidth}p{0.34\textwidth}p{0.34\textwidth}@{}}
\toprule
Expected class & Target statement & Coherent control \\
\midrule
Contradiction & A plane figure is both a perfect square and a perfect circle in the same geometric respect. & A square is drawn inside a circle. \\
Contradiction & The object is entirely red and not red at the same time and in the same respect. & The object is red on one side and not red on the other. \\
Paradox & This sentence is false. & The previous sentence is false. \\
Paradox & A barber shaves all and only villagers who do not shave themselves. & A barber shaves villagers who request a shave but does not shave himself. \\
Underdetermined & Every plank of a ship is replaced, and the result is claimed to be numerically identical to the original. & Every plank is repainted, and the ship remains the same ship. \\
\botrule
\end{tabular}
\end{table}

The four expected labels were defined in the prompt. \emph{Coherent} denoted a logically and conceptually consistent situation. \emph{Contradiction} denoted jointly asserted conditions that cannot all hold in the stated context and respect. \emph{Paradox} denoted an apparent contradiction generated through self-reference, circularity, infinity, or a rule system. \emph{Underdetermined} denoted a genuine conceptual question that does not itself entail contradiction. The exact instruction was: ``Classify the statement using exactly one label: coherent, contradiction, paradox, or underdetermined. Then explain the classification in one sentence.''

\subsection{Model inference and activation extraction}

Experiments used the instruction-tuned \texttt{google/gemma-3-4b-it} checkpoint \cite{gemma3report}. Gemma 3 4B contains 34 transformer layers with residual width 2,560. The model was loaded from local safetensors with Hugging Face Transformers and evaluated in bfloat16 on an Apple M5 Max using the PyTorch MPS backend. Generation was greedy with a maximum of 48 new tokens.

For every prompt, hidden states were requested during a forward pass before generation. From the embedding output and each transformer layer, I retained the vector at the final prompt token---the position immediately preceding the model's first generated token. The resulting array had shape $85\times35\times2560$ and was stored as float16. Model answers and parsed labels were stored separately. The label parser selected the first standalone occurrence of one of the four permitted labels.

\subsection{Behavioral and linear-probe analysis}

Four-way accuracy was computed by exact label match. Wilson 95\% intervals were used for descriptive binomial error bars. For layer-wise probing, expected labels were collapsed to coherent versus noncoherent; the same collapse was applied separately to the model's predicted labels. At every depth, a logistic classifier
\begin{equation}
\hat{y}=\sigma(\mathbf{w}^{\top}\mathbf{h}_{\ell}+b)
\end{equation}
was trained after feature-wise standardization. Scikit-learn's L2-regularized logistic regression was used with $C=0.1$, class-balanced weights, and a fixed random seed. Five-fold GroupKFold cross-validation grouped examples by the 17 seed identifiers. Balanced accuracy was averaged across folds, and the plotted band is the standard error of those five fold scores. Controls used input token count or TF--IDF features from the statement alone; the latter used word 1--2-grams or within-word-boundary character 3--5-grams, with each vectorizer fitted only on its training fold. At the observed peak layer, labels were shuffled within each family for 1,999 permutations, preserving one coherent and four noncoherent examples per family. The resulting $P$ value was Bonferroni-corrected for the 35 tested representation depths.

\subsection{Principal-component analysis}

PCA was fitted independently to the unstandardized, mean-centered residual states at layers 0, 12, 18, and 33. The two leading components were used only for visualization. At layer 12, I additionally correlated PC1 with residual-vector norm and repeated PCA after unit-normalizing each vector as a scale diagnostic. No clustering statistic was derived from the PCA projections.

\subsection{Sparse-autoencoder analysis}

Layer-12 states were encoded with the official Gemma Scope 2 residual-post SAE \texttt{gemma-scope-2-4b-it-res-all/layer\_12\_width\_16k\_l0\_small} \cite{mcdougall2025}. The checkpoint is a JumpReLU SAE trained for the Gemma 3 4B IT residual stream and maps $\mathbb{R}^{2560}$ to 16,384 sparse non-negative features. Encoding used SAELens 6.47.1.

Candidate features for Fig.~\ref{fig:sae} were selected descriptively. For each expected class, I computed mean $\log(1+z_j)$ for every feature $j$, standardized each feature's four class means, and required the target class to have the maximum mean. Features also had to fire on at least 5\% of target-class examples. The two highest-scoring unique features per class were plotted, and their log-activations were correlated with layer-12 residual norm. This is an in-sample exploration; no inferential $P$ values are reported.

\subsection{Software and reproducibility}

The analysis used Python 3.11, PyTorch 2.13.0, Transformers 5.14.1, NumPy 2.4.6, scikit-learn 1.9.0, Matplotlib 3.11.1, and SAELens 6.47.1. Stimuli, extraction code, SAE analysis, exact run configuration, figure data, and plotting code are retained with the project. The deterministic extraction was repeated once and produced byte-identical response and activation files. No language-model weights were modified.

\backmatter

\bmhead{Data availability}

The complete stimulus JSON, model responses, expected labels, and derived figure statistics will be deposited with the public code repository before arXiv submission. Model weights and SAE weights are available from their respective Hugging Face repositories. % TODO: add permanent repository URL.

\bmhead{Code availability}

All scripts required to reproduce activation extraction, SAE encoding, and figures will be released at \url{https://github.com/USERNAME/REPOSITORY}. % TODO: replace URL.

\bmhead{Acknowledgements}

This work was conducted as an independent exploratory project, unaffiliated with and unfunded by any research program.

\bmhead{Author contributions}

The author conceived the study, constructed the stimuli, implemented and ran the analyses, interpreted the results, and wrote the manuscript.

\bmhead{Competing interests}

The author declares no competing interests.

\setlength{\bibsep}{2pt}
\bibliography{references}

\end{document}